# IS THE FUTURE OF AI GREEN?
# WHAT CAN INNOVATION DIFFUSION MODELS SAY ABOUT GENERATIVE AI'S ENVIRONMENTAL IMPACT?

**Robert Viseur**
Université de Mons,
`Robert.Viseur@umons.ac.be`

**Nicolas Jullien**
IMT Atlantique, LEGO-Marsouin,
`Nicolas.Jullien@imt-atlantique.fr`

March 13, 2026

## ABSTRACT

The rise of generative artificial intelligence (GAI) has led to alarming predictions about its environmental impact. However, these predictions often overlook the fact that the diffusion of innovation is accompanied by the evolution of products and the optimization of their performance, primarily for economic reasons. This can also reduce their environmental impact. By analyzing the GAI ecosystem using the classic A-U innovation diffusion model, we can forecast this industry's structure and how its environmental impact will evolve. While GAI will never be green, its impact may not be as problematic as is sometimes claimed. However, this depends on which business model becomes dominant.

The development of artificial intelligence (AI) raises concerns about its environmental impact. A global coalition of over 50 prominent climate and anti-disinformation organizations warned of a potential doubling of data centers, which could result in an 80% increase in global carbon emissions from these infrastructures[1]. The industry promotes the production of decarbonized energy, especially nuclear power, as a solution[2].

These concerns apply particularly to generative artificial intelligence (GAI). GAI refers to "computational techniques that are capable of generating seemingly novel, meaningful content such as text, images, or audio from training data" [1]. They develop solutions (e.g. ChatGPT) based on an underlying model (e.g. GPT or LLaMA) [1]. Large Language Models (LLMs) are an example of such models and can be used to generate text. Training the GPT-3 Large Language Model (LLM) would have required 1.3 GWh, which is equivalent to the average annual energy consumption of 120 U.S. households, and the emission of 522 tons of carbon dioxide [2, 1]. The cost of training these models would double every few months [3]. Using these solutions is also very impacting. A single ChatGPT query would generate "100 times more carbon than a typical Google search" [2].

However, does it make sense to forecast the impact of an emerging technology like GAI to compare it to a mature technology like Google Search?

Economics can help answer this question. We argue that the diffusion of innovation model, known as the "A-U model" after its authors [4], helps us better analyze the evolution of GAI technologies, viewed as software solutions, in terms of its business model and environmental impact.

Technical innovation follows an industrial life cycle that can be divided into three phases [4]. The first phase (A-U:1), or the "fluid" phase, involves an explosion of different product and service designs from various competitors.

Eventually designs consolidate around "*(a) dominant design(s)*" that is/are adopted by all or most of the players in the market. A dominant design is built as an efficient response to user needs, and to producers' needs to build sustainable business models. This second phase (A-U:2) called the "*transitional phase*", is the phase of a process innovation race in which players try to optimize their processes by improving the quality and reducing the cost of their products and services.

---

[1]"Disinformation. Artificial intelligence threats to climate change," CAAD, 2024. Accessed February 2025

[2]"Sam Altman's Fusion Power Startup Is Eyeing Trump's $500 Billion AI Play", D. Jeans, Forbes, February 5, 2025. Accessed February 2025.

Finally, as the processes mature, they gradually become simpler and significantly less expensive. In the "*specific phase*", or A-U:3, the number of players in the market decreases as those who can no longer compete move on to other products, and innovation focuses on incremental changes. The product gradually becomes a commodity while innovation reemerges in new market niches.

What is the current state of GAI?
Regarding the product, – the "interaction interface" [1] offered to users – a dominant design seems to have gradually emerged: the generative conversational agent. However, its precise functionalities have not yet stabilized. Other designs, such as generative meta-engines and APIs, are competing for user acceptance. Consequently, generative AI (GAI) can be considered to be at the end of the A-U:1 phase, as only a few designs remain, yet innovation remains strong. We will show that these designs mirror classic software solution designs and lay the groundwork for anticipating the prevalent designs (A-U:2) and the potential optimizations that could benefit the environment (A-U:3).

Accordingly, the article is structured by the three stages of the technical innovation diffusion model. First, we will present the different current strategies that are typical of the end of an A-U:1 phase. Section 2 will discuss the possible dominant designs that could emerge in the A-U:2 phase. Section 3 will discuss their environmental impact and thus which types of designs could be the most sustainable in A-U:3 phase.

## 1 Current Phase (end of A-U:1): Technical innovation drives impact reduction

GLaM, a model that appeared 18 months after GPT-3, claimed to have achieved a threefold reduction in energy consumption [5]. An evaluation of 400 LLMs showed that their computing resource requirements were halved every eight months or so [6]. In a recent report [7] Google claimed a "44x reduction in carbon footprint for the median Gemini Apps text prompt over the past year". How has this happened?

The International Energy Agency provides a breakdown of energy consumption by artificial intelligence[3]. Up to 10% is spent on model development (experimentation), 20-40% on training, and 60-70% on inference (using the model in production).

While training remains dependent on the availability and advancement of GPUs, inference has seen the development of dedicated chips, such as Groq's LPU (Language Processing Unit). These chips are characterized by faster response times and greater efficiency[4]. Second, training methods are improving (e.g., early stopping, sparse training, and gradient accumulation). Third, the models themselves are optimized. This involves reducing the weight of nodes within the neural network and applying the MoE (Mixture of Experts) principle, as well as reducing model size[5]. These models require fewer computing resources, resulting in a smaller environmental footprint, for comparable performance [8], [9].

Although innovations in models and execution platforms tend to reduce the environmental impact of solutions, they do not yet provide a clear forecast of their future impact.

Examining the offerings developed by the industry through the lens of software economics helps predict the dominant designs that will emerge in Phase 2 and clarify their potential environmental impact.

## 2 Forecasting the Dominant Designs of A-U:2

GAI develops software-based solution, and the economics of such solutions has been explained for quite some time. A software solution and its producer must balance somewhat contradictory goals [10]: to be as adaptable and "customizable" as possible, because of the variety of requirements, yet also as generic and stable as possible, to be resilient in the long run (without too much evolution) and benefit from economies of scale (by dividing the cost of development among as many users as possible).

A producer can achieve this by creating a solution and selling it to as many customers as possible. The goal is to create a standard product that can address the largest possible population with the least amount of adaptation in order to benefit from economies of scale. This is how we understand the promises and positioning of large platforms, such as ChatGPT.

---

[3]"Data centres and data transmission networks", International Energy Agency, accessed Feb. 25.
[4]""Groq Demonstrates Fast LLMs on 4-Year-Old Silicon", S. Ward-Foxton, EETimes, September 12, 2023. Accessed February 2025.
[5]When ai's large language models shrink", by E. Gent IEEE Spectrum, March 31 2023. Accessed February 2025.

Table 1: A summary of information regarding OpenAI's revenue in 2024.

| Fixed costs: | R&D and training | $5B |
|---|---|---|
| | Wages: | $1.5B |
| Variable costs: | Inference | $2B |
| Users & Revenue | The average number of weekly active users (WAU) in 2024* | 200M |
| | Conversion rate to the paying model: | 5% |
| | Monthly revenue per converted user: | $20 |
| | subscriptions' share of revenue: | 70% |

* WAU is the average number of unique users who engaged with (here) OpenAI's services at least once within a specific seven-day period in 2024. See Wikipedia's article for an explanation of why this measure is used to estimate website audience size.

A consortium of users can also develop a solution that meets common or complementary needs. The cost per user may be higher because fewer people share the solution. However, the advantage is that it is closer to individual needs. This is how we envision the development of smaller, more targeted GAI based on specific data.

Various intermediate arrangements are possible. Actors may hire a vendor to develop a solution for them, or a vendor may create a specific offering for a targeted audience. However, the two classic extremes in terms of technical and economic interest are large, unfocused platforms and dedicated solutions. Thus, our analysis will focus on these two models, detailing how they create and capture value, as well as their environmental impact.

## 2.1 Large platform model

The large platform model aims to sell the same service to as many users as possible. This model is efficient because developing software solutions is primarily a fixed-cost endeavor; therefore, sharing a solution creates significant economies of scale.

The GAI industry is well positioned to generate and benefit from economies of scale. Training and development costs remain constant regardless of the user base, and centralizing infrastructures reduces the cost associated with each query.

We evaluated the impact of OpenAI's user base growth on its revenues using published information about its 2024 business and portfolio of models[6]. This portfolio relies primarily on a general-purpose model (e.g., GPT-4o) and a flagship reasoning-capable model (e.g., o1) that is invoked either on demand by the user (if their plan allows it) or automatically based on optimization constraints (e.g., via a real-time router starting with version 5 of the model). Table 1 presents the informations based on 2024's revenues.

Assuming a 5% conversion rate to the paying model, we calculated the estimated cost per user (including estimated production cost and inference cost by user), the revenues and the costs for an increasing number of users. This simplified model allows us to roughly estimate the level of demand needed for the model to be profitable (see Table 2).

First, due to the importance of fixed costs (e.g., training costs), a large installed bases is required. It is the most significant in terms of cost per user until it reaches more than 500 million WAU. As training costs for the most powerful models increase[7], it is difficult to decrease this threshold. Optimizing inference is also imperative for larger competitors because they have the largest installed base. Additionally, multimodal models process more demanding tasks, such as those involving images or videos.

Second, the freemium model is showing its limitations. Our simulation estimates that OpenAI requires a WAU base of over 900 million to break even financially. Competition (OpenAI, Gemini, Microsoft, Mistral, Claude, DeepSeek-AI...) makes it difficult to reduce the benefits of the free version, but also to develop this base. The free access is a classic in a business model based on selling the attention of users, notably through advertising, imitating the model of Google or

---

[6] Precisely the articles by Nickie Louise "OpenAI bleeding money" and by George Hammond and Cristina Criddle "OpenAI makes 5-year business plan to meet $1trln spending pledges"", retrieved in October 2025. The statistics on ChatGPT usage come from the report "How people use ChatGPT".

[7] See figures related to GPT-3 and GTP-4 models.

Table 2: A simulation of OpenAI's revenues and costs according to the number of monthly users (USD).

| Weekly active users (WAU) | Fixed costs per WAU* | Variable costs per WAU* | Total costs | Total revenues |
|---|---|---|---|---|
| 200M | 32.5 | 10 | 8.5B | 3.4B |
| 500M | 13.0 | 10 | 11.5B | 8.6B |
| 700M | 9.3 | 10 | 13.5B | 12.0B |
| 900M | 7.2 | 10 | 15.5B | 15.4B |
| 1.0B | 6.5 | 10 | 16.5B | 17.1B |

* dividing the total costs of Table 1 by the number of WAU. The inference cost is probably decreasing slightly with the number of users. The optimization and innovation efforts mentioned in the article may decrease this cost in the future as well.

* In calculating the revenues, we assumed that the WAU was close to the number of monthly active users because active users are active every week.

Facebook. The fact that chatbots are now being used massively by younger people, as research tools [11], is a strong indication that these two markets (GAI and research engine) may converge in the future.

In an audience race, performance announcements are as much aimed at advance customers as they are at occupying the news to attract customers. The DeepSeek-R1 LLM model illustrates how this race is refocusing on process innovation rather than the growth of the model. Developed by the Chinese company DeepSeek-AI, it was released against the backdrop of growing tensions and competition between the US and China. The model is notable for its MoE architecture, which claims to reduce both training and inference costs [8] with performance comparable to the best US models.

In summary, the large platform model is characterized by a financial and environmental race for market share due to economies of scale and the fact that audience leadership attracts a dominant share of subscriptions or advertising revenues. Only a few companies will survive using this model. This is probably better for the environment because each company will optimize its infrastructure and usage.

But paid usage is also gradually shifting to complex tasks that require models capable of reasoning (e.g., OpenAI o1) or automating the processing of more documents (e.g., deep search capabilities).

The goal behind this shift is to integrate these tools into companies' information system workflows and services. This is already the case with features like writing assistance on LinkedIn and automatic summarization of articles in generative search engines like Perplexity or Elicit.

Microsoft is a case in point. Microsoft's Copilot, based on GPT-4 and 5, is integrated into its various software solutions, including productivity software and chatbot solutions. The company justifies and finances this integration by increasing subscription costs for these solutions[8].

However, addressing specific business domains using large platform models requires developing RAG (retrieval-augmented generation) solutions. RAG improves response accuracy by using up-to-date information and domain-specific knowledge without requiring LLM specialization. To be truly effective, models must thus be fine-tuned with specific prompts and extensive or dedicated datasets. This has led some actors to adopt a different strategy. Rather than tuning afterward the best basic generative model, they are exploring the development of smaller, dedicated models.

## 2.2 The Targeted Model Strategy

An evaluation of 70 of these small language models (SLMs) [9] stressed that they innovate in several directions, such as dataset and model organization. It makes them a competitive alternative to large language models (LLMs) when focusing on specific tasks. Due to their lower deployment costs, they are also more diverse.

The simplest business model for these specialized GAIs is to develop and integrate them into existing solutions. This is the strategy followed by Amazon with its conversational agent Rufus. Rufus allows users to ask shopping-related questions on Amazon's platforms. Rufus's language model is trained on data from product catalogs as well as customer questions and reviews[9].

---

[8]"Microsoft 365 sees 43% price hike thanks to Copilot - existing customers safe until renewal", by Christopher Harper, Tom's Hardware, published January 17, 2025, accessed February 2025.

[9]The design and optimization of Rufus have been described by Amazon's teams in IEEE Spectrum.

This targeted model strategy may also be supported by the sovereignty policies promoted by certain states, aimed in particular at defending their native language, developing their own technological capabilities or mitigating the infrastructural power of global platforms [12].

However, this model comes at the cost of specialization in uses and, consequently, the market served [13].

These considerations highlight two complementary approaches to developing GAI-based tools. The first uses large, generalist models, and the second uses smaller, more specialized models integrated into a broader toolkit to address specific needs. The first approach lends itself well to economies of scale and winner-take-all strategies. The second approach has a lower barrier to entry and lends itself to more diverse applications. This could result in a proliferation of models dedicated to niche markets.

The next and final session will discuss the consequences for GHG emissions if these designs become dominant.

## 3 Models' GHG Emissions and beyond (A-U:3)

### 3.1 Models impacts

Large, general-purpose models (e.g., GPT, Gemini, Claude, and DeepSeek) are characterized by their significant GHG emissions. As use gradually shifts towards complex tasks requiring models capable of reasoning (e.g., OpenAI o1) or automating the processing of more documents (e.g., deep search functions), the impact will remain strong because the larger the model, the greater the energy cost during inference [14]. As this market mainly focuses on audience share, it will stabilize around a few optimized solutions, dramatically reducing the total environmental impact of these models. However, competition for audience share will encourage model owners to maximize use of these models, thus increasing their environmental impact.

The smaller models (e.g., LlaMA 7B, Rufus, or Pleias) are characterized by significantly lower emissions for comparable, if not better, accuracy. However, if each model requires less training, more models will be needed. The risk is that a greater number of solutions with a smaller environmental impact could result in a greater total impact.

As with any software solution, the challenge is to strike the right balance between a shared, standardized solution and personalization. As with other software, openness may be helpful.

### 3.2 Open GAI as a solution?

Opening the models facilitates local optimization of shared models, as illustrated by the Sonar large language model family build by Perplexity. Based on the open Llama model family, Sonar "is optimized for answer quality and user experience"[10]. The model's specialization enables it to achieve high processing speeds and, consequently, lower inference costs.

However, it also highlights the limitations of open foundation models. Perplexity integrates Llama into a closed, cloud-based model, but it's unclear if this improves Llama. How can the cost of developing and maintaining the foundation model be financed?

Public support can help. The French LUCIE LLM project, has been developed by a consortium of actors in France, OpenLLM France coordinated by a open-source software engineering company, Linagora. It has been trained in a shared, existing public infrastructure, the Jean Zay supercomputer, reducing costs and environmental impact.

Secondly, training a model requires a lot of feedback, usually provided by "click workers". As stated on LUCIE's website, or on Meta's LLama 3 website, opening the models and tests helps obtain direct feedback on tasks that are useful for users and potentially more specific datasets. It also helps to explore some optimization solutions as long as users publish them back.

For service companies, such as Pleias, adopting an open strategy with dedicated foundation models allows users to test them. Those users may become clients when they need to adapt these models to their needs and data, for which they will require expertise. This approach also demonstrates good faith, as it ensures users will always have access to the model. The risk is that another actor could propose a competing service based on this model. However, Pleias and LUCIE's experiences show that this threat can be reduced by segmenting markets and targeting specific objectives for certain sectors. This approach may result in low reusability.

Two types of players seem particularly interested in such tools openness: 1) service companies whose business model is based on adapting existing solutions to the specific needs of their customers (e.g. Linagora, which participates in the development of LUCIE); 2) large software users who want to retain control over their solutions (for example, to remain in the French context, companies have created the TOSIT structure to collaborate on shared FOSS projects, including AI projects such as the Tock open conversation kit). This second category corresponds to the user innovators at the origins of FOSS [15]. The first category sees an opportunity to serve these clients. For these actors, added value comes from optimizing tools for training and fine-tuning, as well as tools for optimal exploitation of these models (e.g., RAG software). Since these training and inference methods are closely tied to the core model, these actors may be incentivized to invest in it, to ensure compatibility, but also to reduce the risk of lock-in, one motivation for the adoption of FOSS [15], since the model can be rebuilt in case of disagreement with the main branch. Instead of maximizing its market value, a solution managed by a consortium of users is incentivized to attract as many users as possible in order to decrease its cost per user. This would decrease the number of models and optimize the global environmental impact of GAI.

---

[10]See the description of Sonar model.

## 4 Conclusion

The cycle described by the A-U model was used to analyze the current development of GAI and forecast the evolution of the industry and its greenhouse gas (GHG) impact. Generally, diffusion during an innovation process leads to optimization and reduction of individual consumption, which is already occurring in GAI.

Pioneer models have been published for simple tasks (e.g. GPT-3.5). These models, created by bigtechs, have demonstrated their usefulness and the plebiscite for conversational agent design (e.g. ChatGPT), but also their limitations for more complex tasks, such as reasoning chains, and a certain inefficiency in terms of resource consumption. This has led to a dual pressure for innovation. On the one hand, the limitations have encouraged the introduction of more powerful and heavier models (e.g. OpenAI o1). On the other hand, the existence of a user base for simpler models (e.g. OpenAI GPT-4o) has encouraged their optimization. Continued product innovation based on new models has been complemented by process innovation aimed at reducing production costs (new training strategies) and operating costs (optimization of data centers; agents, specialized chips, etc.) while maintaining performance (e.g. DeepSeek-V3 and DeepSeek-R1 [8]).

This will eventually decrease the impact per user of GAI. However, the competition for audience share among major models or the potential proliferation of smaller models could jeopardize these efforts by increasing their usage and, consequently, their global impact.

In this article we advocate that the FOSS model exemplified by LUCIE-7B or Pleias, or the Linux Foundation's AI & Data initiative may be one way to address this conundrum, because it helps limit the proliferation of models, and increase the sharing of best practices.

However, more innovation and commercial exploration is needed to confirm its economic sustainability. But from what research can tell [15], in terms of the FOSS business model, FOSS AI seems quite classic: digital services companies and key user innovators aim to develop generative business solutions and benefit from cumulative efforts to optimize model training and use. This model does not focus solely on the model, but rather on the entire ecosystem, including the model, training software, derived tools, and datasets. Players are free to adapt solutions (such as re-training on specific data and tasks), but they must collaborate on the foundation model to ensure that their innovative offerings remain compatible with it.

If this design were to become dominant, controlling GHG emissions would come from two sources. First, the models used would be simple. Second, a limited number of FOSS foundation models could emerge, which would reduce or even eliminate the rebound effect. These changes would complement advances made by bigtech in specialized hardware and public cloud hosting.